\documentclass[conference]{IEEEtran}
\IEEEoverridecommandlockouts
\usepackage{cite}
\usepackage{amsmath,amssymb,amsfonts}
\usepackage{algorithmic}
\usepackage{graphicx}
\usepackage{textcomp}
\usepackage{xcolor}
\usepackage{adjustbox}
\usepackage{bm}
\usepackage{booktabs}
\usepackage{threeparttable}

\usepackage[caption=false,font=footnotesize]{subfig}

\def\BibTeX{{\rm B\kern-.05em{\sc i\kern-.025em b}\kern-.08em
    T\kern-.1667em\lower.7ex\hbox{E}\kern-.125emX}}
\begin{document}

\title{GraD-IBD: Graph Representation Learning from Diagnosis Trajectories for Early Detection of Inflammatory Bowel Disease}

\author{
    \IEEEauthorblockN{
        Leo Y. Li-Han\IEEEauthorrefmark{1},
        Ellen L. Larson\IEEEauthorrefmark{1},
        Elizabeth B. Habermann\IEEEauthorrefmark{2}, 
        Cornelius A. Thiels\IEEEauthorrefmark{3},\\       
        Hojjat Salehinejad\IEEEauthorrefmark{2}\IEEEauthorrefmark{4}, Senior Member, IEEE
    }
    \IEEEauthorblockA{
        \IEEEauthorrefmark{1}Department of Surgery, Mayo Clinic, Rochester, MN, USA\\
        \IEEEauthorrefmark{2}Kern Center for the Science of Health Care Delivery, Mayo Clinic, Rochester, MN, USA\\
        \IEEEauthorrefmark{3}Division of Hepatobiliary and Pancreas Surgery, Mayo Clinic, Rochester, MN, USA\\
        \IEEEauthorrefmark{4}Department of Artificial Intelligence and Informatics, Mayo Clinic, Rochester, MN, USA
    }
    \IEEEauthorblockA{
        \begin{minipage}{\linewidth}\centering
        \texttt{\{li-han.leo, larson.ellen1,
        habermann.elizabeth, thiels.cornelius, salehinejad.hojjat\}@mayo.edu}
        \end{minipage}
    }

\thanks{This paper was accepted for presentation at the 2026 IEEE Engineering in Medicine and Biology Society (EMBC), Toronto, Canada, July 2026}
}

\maketitle

\begin{abstract}
International Classification of Diseases (ICD) is a globally recognized coding system that records diagnostic events during each patient encounter, providing a standardized data foundation for various clinical tasks. However, the irregular and hierarchical nature of ICD code sequences poses challenges for $N$-D lattice-based sequential modeling methods, leading to overly complex model designs. In this paper, we propose \textit{GraD-IBD}, a \textit{gra}ph \textit{d}iagnosis model that reformulates longitudinal ICD trajectories as visit-bucketized, temporally directed graphs to detect the risk of inflammatory bowel disease (\textit{IBD}). A novel context-aware, time-decay message passing mechanism was developed to capture temporal dependencies while reducing model complexity. The experimental results using a real-world clinical dataset demonstrated consistent and robust improvements in IBD detection over state-of-the-art methods, with significant reductions in computational complexity compared to sequential models. These findings highlight the potential of graph representation learning to enable efficient, scalable, and accurate disease risk prediction from longitudinal ICD diagnosis codes.
\end{abstract}


\section{Introduction}
The International Classification of Diseases (ICD) is a widely adopted alphanumeric coding standard for categorizing and recording diagnostic events in every healthcare visit~\cite{world2009international}. Longitudinal ICD code time series provide a compact and informative representation of patient health trajectories, making them a crucial data source for various clinical tasks, including disease onset detection~\cite{li2020behrt}, readmission prediction~\cite{rajkomar2018scalable}, and medication recommendation~\cite{junyuan2019pre}. 

Inflammatory bowel disease (IBD) refers to a group of conditions that cause chronic inflammation of the gastrointestinal tract~\cite{baumgart2007inflammatory}. As IBD manifests in various nonspecific symptoms, such as abdominal pain, bleeding, and diarrhea~\cite{perler2019presenting}, it is challenging to diagnose in sporadic non-specialty healthcare settings. The average delay from symptom onset to IBD diagnosis is 3-6 months, leading to late treatment and compromised outcomes, such as higher rates of surgical intervention and complications~\cite{jayasooriya2023systematic}. Adults under age $50$ in the U.S. are at high risk of delayed diagnosis due to lower rates of healthcare utilization, insurance barriers, and limited availability of definitive testing using colonoscopy in this population~\cite{CDCstatsWellnessVisit2024, CDCNoInsurance2021}. Therefore, identifying the risk of IBD based solely on diagnosis trajectories is particularly valuable for this population whose medical records are typically sparse.

\begin{figure}[t]
\centerline{
    \includegraphics[width=1\linewidth]{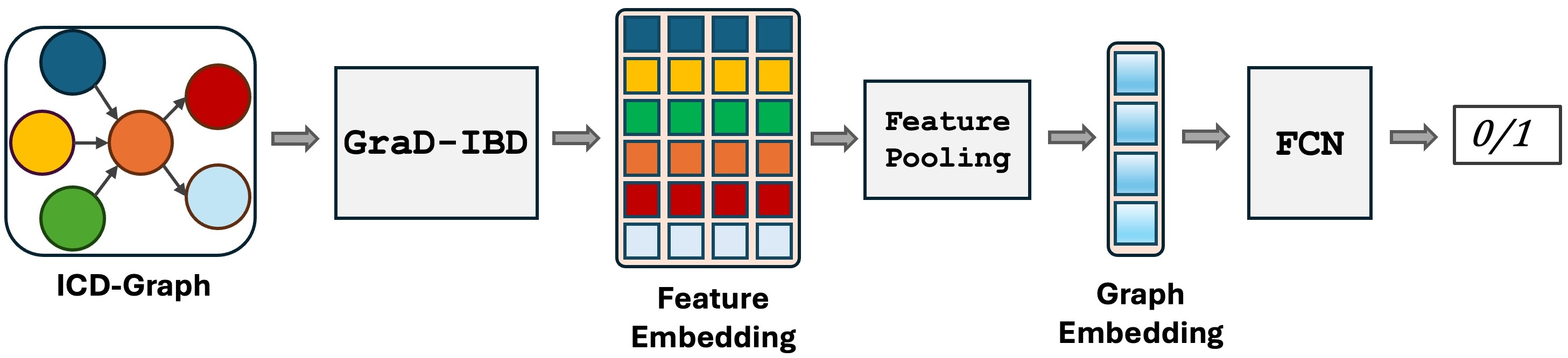}
}
\caption{The overall architecture of the proposed Grad-IBD model. Unstructured longitudinal ICD codes per patient are reformulated into a concise graph representation (ICD-graph). A graph neural network (GraD-IBD) equipped with a context-aware, time-decay message-passing mechanism learns graph features for disease risk detection. }
\label{fig_architecture}
\end{figure}

Disease detection can be formulated as a time-series classification problem, where a patient's longitudinal medical history, combined with static variables, serves as input to a classifier~\cite{salehinejad2023contrastive,salehinejad2023novel,salehinejad2025deep}. Graph representation learning has attracted increasing attention in bioinformatics and medical time series analysis due to its unique ability to model complex data structures and features~\cite{choudhary2024graph}. 
For instance, Choi~\textit{et al.}~\cite{choi2017gram} introduced an attention-enabled medical ontology graph to facilitate learning data embeddings in disease onset detection. Similarly, Jiang~\textit{et al.}~\cite{jiang2023graphcare} incorporated diverse medical concepts to construct personalized knowledge graphs, resulting in comprehensive data representation and improved performance across various predictive tasks. In other efforts~\cite{choi2020learning, wen2022disentangled, chen2024predictive}, assorted, heterogeneous, and temporal graphs have been proposed, respectively, to learn spatio-temporal associations to boost prediction performance. Considering IBD's complex and diverse symptom evolution patterns, this underscores the need for effective models to disentangle diagnostic trajectories from those of other similar diseases.

In this study, we developed a streamlined graph-based diagnosis model using ICD diagnosis code trajectories for IBD risk triage. To the best of our knowledge, this is the first to leverage graph representation learning in IBD management. The contributions of this work are threefold.
\begin{itemize}
    \item We introduce a \textbf{visit-bucketized frequency encoding paradigm} for irregular ICD code time series.
    \item We design a \textbf{patient-level, temporally directed graph} (\textbf{ICD-graph}) to reformulate diagnosis encodings.
    \item We build \textbf{the GraD-IBD model} with a novel context-aware and time-decay message passing mechanism to learn meaningful features from each ICD-graph.
\end{itemize}

\begin{figure}[!t]
\centerline{
    \includegraphics[width=1\linewidth]{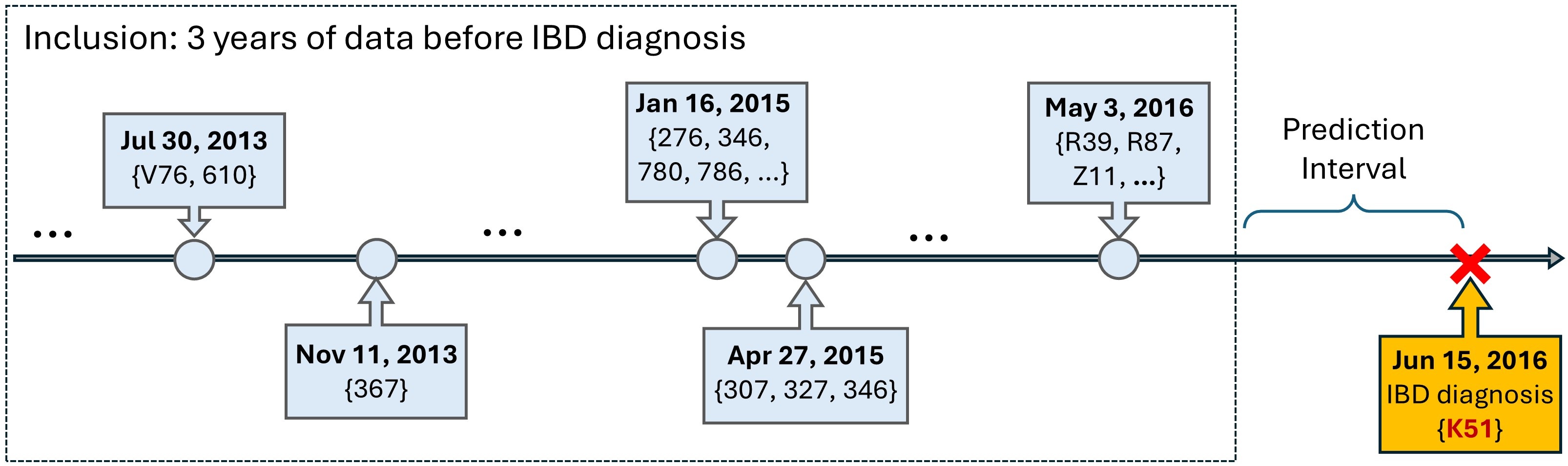}
}
\caption{Sequential representation of a patient's ICD diagnostic history. The task aims to determine whether a patient will develop IBD with a lead time (i.e., prediction interval), based on ICD codes received in the 3 years preceding the diagnosis of interest (yellow box). }
\label{fig_icd_seq}
\end{figure}

\section{Methods}
The proposed GraD-IBD model addresses the challenges of modeling irregular and unstructured ICD code sequences by reformulating patient diagnosis histories into structured graph representations. It follows three main steps: (i) encoding longitudinal ICD sequences into visit-bucketized representations, (ii) constructing temporally directed patient-level graphs with weighted edges that capture diagnostic co-occurrence and temporal dependency, and (iii) developing a graph neural network with context-aware, time-decay message passing to learn discriminative graph embeddings for IBD risk prediction. This design enables efficient modeling of temporal dependencies while maintaining low computational complexity compared to conventional sequence models. The overall architecture of the proposed method is illustrated in Fig.~\ref{fig_architecture}.

\subsection{ICD Codes Encoding}\label{encoding}
As shown in Fig.~\ref{fig_icd_seq}, a patient's diagnostic history consists of a sequence of irregularly spaced clinical visits, each containing a variable number of ICD diagnosis codes. For notational simplicity, we focus on the data representation for a single patient, thereby ignoring patient-specific notation in the following expressions. 

Let $\mathcal{D}=\{d_1, d_2, ..., d_{{\mathrm{N}}}\}$ be a set of unique ICD diagnosis codes in our dataset, and $\mathcal{C}= (1, 2, ..., \mathrm{N})$ be the code indices, where $|\mathcal{D}|=\mathrm{N}$ and $|\cdot|$ is the cardinality. 
We first introduced a visit-bucketized frequency encoding paradigm to translate the irregular ICD sequence per-patient into a structured matrix $\mathbf{X}\in\mathbb{R}^{\mathrm{N}\times \mathrm{T}}$, where $\mathrm{T}$ is the total number of buckets, with each bucket representing a duration of $\tau$ days, and frequency is defined as the number of times each diagnosis code appears in a bucket. When $\tau=1$, our bucketization is equivalent to the per-visit data formation in previous studies~\cite{chen2024predictive}. Let $\mathcal{T}=( 1,...,\mathrm{T})$ be the set of bucket (column) indices of $\mathbf{X}$. The value of each element $x_{c,t}$ of $\mathbf{X}$ is the occurrence frequency of diagnosis code $c$ observed in bucket $t\in\mathcal{T}$.

The clinical assumption for grouping visits into buckets is that diagnoses received within a short period, such as a week, can serve as a holistic snapshot of the patient's status at that moment, rather than exhibiting separate temporal or causal dependencies. As such, the visit bucketization design would help simplify data representation and modeling complexity.

\begin{figure}[!t]
\centering
\includegraphics[width=0.88\linewidth]{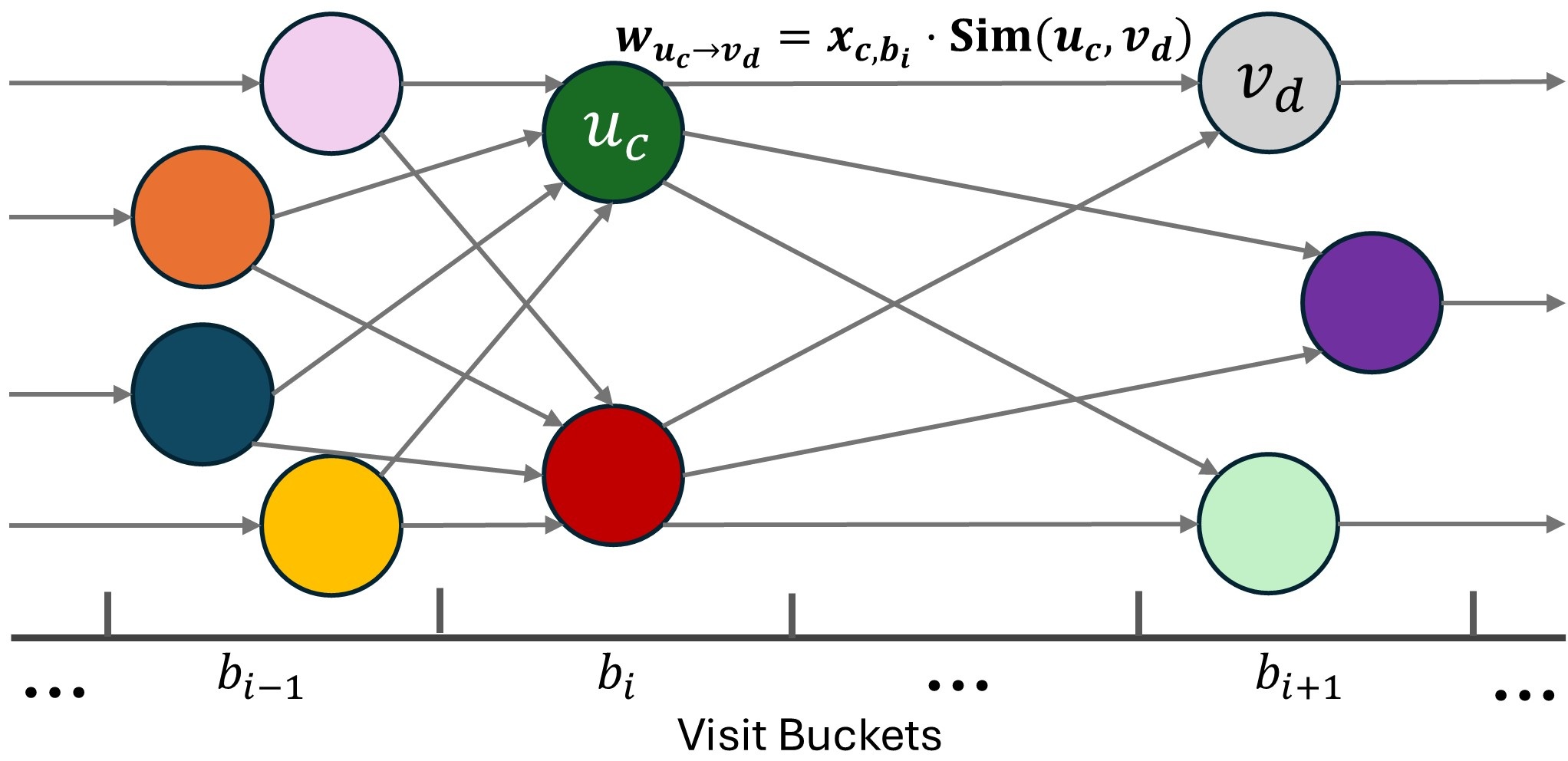}
\caption{Illustration of the ICD-graph for a patient. Nodes represent unique ICD codes observed in each chronologically ordered visit bucket. Node features correspond to ICD code embeddings defined by a learnable embedding matrix. Edges are directed from each node in one bucket ($b_{i}$) to all nodes in the subsequent non-empty bucket ($b_{i+1}$). Edge weights are defined by the cosine similarity between the source ($u_c$) and target ($v_d$) node embeddings multiplied by the occurrence frequency of code $c$ in bucket $b_{i}$, i.e., $x_{c,b_i}$. }
\label{fig_graph}
\end{figure}

\subsection{ICD-Graph} \label{graph}
By further compressing the visit-bucketized encoding matrix $\mathbf{X}$ along both dimensions, we created a temporally directed graph $\mathcal{G}(\mathcal{V}, \mathcal{E})$ to represent the ICD code sequence for each patient, where $\mathcal{V}$ and $\mathcal{E}$ represent sets of nodes and edges, respectively. The node set $\mathcal{V}$ comprises all ICD codes observed in the patient's diagnostic history. Recurring codes over time are represented by distinct nodes in different visit buckets. 

Specifically, let $\mathcal{B}=( b_1, b_2, ..., b_\mathrm{K} ) \subseteq \mathcal{T}$ be a subset of chronologically ordered non-empty visit buckets in $\mathbf{X}$, where $b_1<b_2<...< b_\mathrm{K}$ and $|\mathcal{B}|=\mathrm{K}\leq \mathrm{T}$. For each $b_i\in\mathcal{B}$, we define a node subset $\mathcal{V}^{(b_i)}$ containing unique codes presented in the bucket, i.e., $\mathcal{V}^{(b_i)}=\{u_c~|~x_{c,b_i}>0, c\in \mathcal{C} \}$. Note that, for notational simplicity, we ignore the visit bucket superscript for node $u_c$ in the text and use $u_c \in \mathcal{V}^{(b_i)}$ instead when needed. Thus, $\mathcal{V}$ is the union of all node subsets: $\mathcal{V} = \bigcup_{b_i \in \mathcal{B}} \mathcal{V}^{(b_i)} $. The number of nodes in $\mathcal{V}$ equals the non-zero elements of $\mathbf{X}$, i.e., $|\mathcal{V}|=\sum_{c\in\mathcal{C}}\sum_{t\in\mathcal{T}}\mathbf{1}\{x_{c,t}>0\}$, where $\mathbf{1}\{\cdot\}$ is the indicator function.

The edge set $\mathcal{E}$ contains fully connected, directed connections from nodes in $\mathcal{V}^{(b_i)}$ to $\mathcal{V}^{(b_{i+1})}$, $\forall \;i\in \{1,...,\mathrm{K}-1 \}$, representing the temporal flow of the patient's diagnostic history, i.e., $\mathcal{E} = \bigcup_{i=1}^{\mathrm{K}-1} (\mathcal{V}^{(b_i)} \bm{\times} \mathcal{V}^{(b_{i+1})}) $, where $\bm{\times}$ is the Cartesian product. Furthermore, we assign a weight to each edge $e=(u_c, v_d) \in \mathcal{E}$ to represent the strength of the connection from node $u_c\in \mathcal{V}^{(b_i)}$ to $v_d \in \mathcal{V}^{(b_{i+1})}$. The edge weight function is defined as
\begin{equation}\label{eq:edge_wt}
    f(e) = x_{c,b_i} \cdot \text{Sim}(u_c, v_d)    ,
\end{equation}
where $x_{c,b_i}$ represents the occurrence frequency of code $c$ in bucket $b_i$ and $\text{Sim}(\cdot)$ measures the cosine similarity between the source $u_c$ and target $v_d$ node embeddings that will be elaborated later. Fig.~\ref{fig_graph} depicts three consecutive non-empty visit buckets in the ICD-graph for a patient.

As a result, each patient's unstructured diagnosis trajectory was converted into a structured and concise graph representation. Accordingly, the IBD detection problem can be framed as a graph-level classification task.

\subsection{Learning on the Graph} \label{gnn}

\subsubsection{Node Feature Embedding}
We developed a graph representation learning model to extract comprehensive embeddings from ICD-graphs, which were then used for IBD risk detection. First, the feature of a node $u_c$ in an ICD-graph was initialized by a learnable embedding vector, denoted $\mathbf{h}_{u_c}\in \mathbb{R}^{d_{\text{node}}}$, which represents the specific ICD code type and corresponds to a row in an embedding matrix $\mathbf{E}\in \mathbb{R}^{\mathrm{N}\times d_{\text{node}}}$. Here, $d_{\text{node}}$ is a hyperparameter defining the node feature dimension. The embedding matrix $\mathbf{E}$ was used across all ICD-graphs in the dataset and was continuously updated during training. 

\vspace{2mm}
\subsubsection{Message Passing, Aggregation, and Feature Update}
Inspired by the GraphSAGE model~\cite{hamilton2017inductive}, we built a novel context-aware, time-decay message-passing mechanism to update the target node's feature vector based on its temporal neighbors from preceding visit buckets. Specifically, for each node $v_d\in \mathcal{V}^{b_{(i+1)}}$, features from incoming nodes $u_c \in \mathcal{V}^{(b_{i})}$, where $c, d \in \mathcal{C}$, are passed and aggregated through a weighted sum defined by the edge weights given in \eqref{eq:edge_wt}. This design ensures that information from similar and repeated diagnoses appearing in the preceding visit bucket carries more weight in message passing to the target node, thereby enabling context-aware learning. Furthermore, weights from all incoming edges are normalized before aggregation, and the normalized edge weights $\tilde{w}_{u_c\to v_d}$ are formulated as 
\begin{equation}
    \tilde{w}_{u_c\to v_d}=\frac{f(u_c, v_d)}{\sum_{u_{c'}\in\mathcal{V}^{(b_{i})}} f(u_{c'}, v_d)}.
\end{equation}
As such, the aggregated message $\mathbf{m}_{\mathcal{N}_d} \in \mathbb{R}^{d_{\text{node}}}$ from all nodes in the preceding bucket $b_{i}$ to node $v_d$ in the current bucket $b_{(i+1)}$ is given by 
\begin{equation}
    \mathbf{m}_{\mathcal{N}_d} = \sum_{u_c\in\mathcal{V}^{(b_{i})}} \tilde{w}_{u_c\to v_d}\cdot \mathbf{h}_{u_c}.
\end{equation}

Next, the aggregated message is used to update the target node's feature vector $\mathbf{h}_{v_d}\in\mathbb{R}^{d_{\text{node}}}$. In this process, the aggregated message $\mathbf{m}_{\mathcal{N}_d}$ is multiplied by an exponential time-decay factor to account for the temporal gap between the two buckets, i.e., information from recent visits weighs more than that from earlier visits. Then, the time-decayed neighbor message is element-wise added to the target node's feature embedding $\mathbf{h}_{v_d}$, followed by a one-layer fully-connected network (FCN) with rectified linear unit (ReLU) activation~\cite{lecun2015deep} to produce the updated node representation $\mathbf{h}_{v_d}^\prime\in \mathbb{R}^{d_\text{graph}}$, where $d_\text{graph}$ is the graph embedding dimension. 

The entire feature updating process for nodes $v_d\in \mathcal{V}^{(b_i)}$ from its immediate neighbors can be expressed as follows
\begin{equation}
    \mathbf{h}_{v_d}^\prime = \text{FCN}\bigl( \mathbf{h}_{v_d} + \mathbf{m}_{\mathcal{N}_d} \cdot \text{exp}(-\lambda\cdot\Delta t_{{i+1}, i}) \bigr),
\end{equation}
where $\lambda$ is a hyperparameter that controls the time decay rate and $\Delta t_{{i+1},i}=b_{i+1} - b_i$ represents the time difference between two consecutive non-empty buckets. In practice, this procedure is recursively applied for ${S}$ times to propagate diagnostic information from earlier buckets, corresponding to the search depth defined in~\cite{hamilton2017inductive}. Therefore, all node embeddings in the graph $\mathcal{G}$ form a matrix $\mathbf{H_{\mathcal{G}}} \in \mathbb{R}^{|\mathcal{V}|\times d_{\text{graph}}}$, where $|\mathcal{V}|$ is the number of nodes in the graph.

\vspace{2mm}

\subsubsection{Feature Pooling and Classification}
As the number of nodes in different graphs can vary, to serve graph-level classification, we applied a global average pooling to the node embedding matrix $\mathbf{H_{\mathcal{G}}}$ across the node dimension (row-wise), resulting in a fixed-length graph embedding vector $\mathbf{z_{\mathcal{G}}}\in \mathbb{R}^{d_{\text{graph}}}$. Finally, a two-layer FCN with Layer Normalization~\cite{ba2016layer} and ReLU nonlinearity was trained to make the binary IBD classification based on the graph embedding $\mathbf{z_{\mathcal{G}}}$.

\section{Experiments on Real-World Data}
\subsection{Data}
We used de-identified clinical diagnosis data retrospectively extracted from the Mayo Clinic Health System, which includes a variety of care settings, ranging from primary care to tertiary referral centers. The diagnostic trajectories of 1167 IBD patients aged 18 to 50 years with ICD codes K50 (Crohn's disease) or K51 (Ulcerative Colitis), who were initially ascribed between October 2015 and January 2024, were included in the case group. Meanwhile, 7871 age and gender matched patients with six different types of diagnostic codes for benign abdominal conditions that mimic common presenting symptoms of IBD were selected as the control group. This case-control selection aims to challenge the model to discover effective features that help distinguish these patient groups with similar symptoms, which can be clinically challenging and valuable.
 
All included patients had 1 to 3 years of diagnosis history prior to the diagnosis of interest. For simplicity, all ICD codes in this study were truncated to the chapter level by keeping the first 3 characters, resulting in $1982$ unique ICD codes plus one \textless{}UNK\textgreater{} token designated for unseen codes (i.e., $\mathrm{N}=1983)$. Before training, data from 903 patients ($10\%$), stratified by the case-control distribution, were held out for final testing. The remaining 8135 patients were used in model training, cross-validation (CV), hyperparameter tuning, and ablation experiments.

\subsection{Training and Evaluation Setup}
All hyperparameters were determined through grid search. For model design, we evaluated multiple bucket sizes $\tau \in \{1, 3, 7, 14, 30\}$ and selected weekly buckets $\tau = 7$ for all experiments. The node feature dimension $d_{\text{node}}$ and graph embedding dimension $d_{\text{graph}}$ were set to $64$ and $256$, respectively, searching over $\{16, 32, 64, 128, 256, 512\}$. The message-passing depth ($S$) and exponential time-decay factor ($\lambda$) were set to $3$ and $0.3$, respectively, chosen from $\{1, 2, 3, 4\}$ and $\{0.1, 0.2, 0.3, 0.4, 0.5\}$. 

During training, we employed 10-fold CV to minimize binary cross-entropy loss using the Adam optimizer~\cite{kingma2014adam}. The training–validation splits also followed the case–control distribution. The initial learning rate was set to $10^{-3}$, selected from $\{10^{-1}, 10^{-2}, 10^{-3}, 10^{-4}\}$, and reduced by a factor of $10$ after three stagnant epochs. The batch size was set to $8$, chosen from $\{1, 4, 8, 16, 32, 64\}$.

In the testing phase, trained models from all CV folds were evaluated on the test set, and performance was measured using the area under the receiver operating characteristic curve (AUROC), Average Precision (AP), and F1 score (with the decision threshold of $0.5$), which assess different performance characteristics of the model. The mean and $95\%$ confidence interval (CI) of those metrics were reported. Additionally, the floating-point operations (FLOP) and the number of learnable parameters of the model were calculated to quantify the computational complexity.

We compared GraD-IBD with different graph-based models, including GraphSAGE~\cite {hamilton2017inductive}, GAT~\cite{velivckovic2017graph}, and GCN~\cite{kipf2016semi}, as well as two sequential modeling methods: Transformer~\cite{vaswani2017attention} and Long Short-Term Memory (LSTM)~\cite{hochreiter1997long}. ICD-graphs were used as input to all baseline graph models, and hyperparameters were chosen from the same sets described above. Note that, as all nodes in a visit bucket may contain valuable information for IBD detection, we employed fully connected GraphSAGE for message passing rather than random neighbor sampling. Additionally, message passing in all graph-based models also followed the directed ICD graph structure. The Transformer model was implemented with the Encoder component described in~\cite{vaswani2017attention}, where the selections of self-attention layers, heads, and dimensions were 1, 16, and 256, respectively. The visit-bucketized encoding matrix was used as model input, and a ``\textless{}CLS\textgreater{}" token was added as the feature vector for classification following the convention in~\cite{devlin2018bert}. The LSTM model took the same input with the layer and hidden dimension of 1 and 256, respectively.

\begin{table}[t]
\centering
\caption{Testing performance and computational complexity of IBD detection models with the one-month prediction interval.}
\label{tab:performance}
\begin{adjustbox}{width=0.49\textwidth}
    \begin{tabular}{@{}cccccc@{}} 
    \toprule
    \textbf{Model}  & \textbf{AUROC}  & \textbf{AP}   & \textbf{F1 score}   & \textbf{\begin{tabular}[c]{@{}c@{}}FLOP\\ (M)\end{tabular}}  & \textbf{\begin{tabular}[c]{@{}c@{}}Params\\ (M)\end{tabular}} \\ \midrule
    \begin{tabular}[c]{@{}c@{}}GraD-IBD\\~\end{tabular}   & \begin{tabular}[c]{@{}c@{}}\textbf{0.752}\\(0.745--0.759)\end{tabular} & \begin{tabular}[c]{@{}c@{}}\textbf{0.597}\\ (0.591--0.604)\end{tabular} & \begin{tabular}[c]{@{}c@{}}\textbf{0.612}\\ (0.604--0.620)\end{tabular} & \textbf{23.495}     & \textbf{0.172}     \\
    \begin{tabular}[c]{@{}c@{}}GraphSAGE\\~\end{tabular}   & \begin{tabular}[c]{@{}c@{}}\textbf{0.752}\\(0.743--0.761)\end{tabular} & \begin{tabular}[c]{@{}c@{}}0.582\\ (0.564--0.599)\end{tabular} & \begin{tabular}[c]{@{}c@{}}0.565\\ (0.544--0.586)\end{tabular} & \textbf{23.495}     & 0.213     \\
    \begin{tabular}[c]{@{}c@{}}GAT\\~\end{tabular}     & \begin{tabular}[c]{@{}c@{}}0.727\\(0.718--0.736)\end{tabular} & \begin{tabular}[c]{@{}c@{}}0.580\\ (0.567--0.592)\end{tabular} & \begin{tabular}[c]{@{}c@{}}0.600\\ (0.580--0.621)\end{tabular} & 28.230     & 0.390     \\
    \begin{tabular}[c]{@{}c@{}}GCN\\~\end{tabular}     & \begin{tabular}[c]{@{}c@{}}0.719\\(0.706--0.733)\end{tabular} & \begin{tabular}[c]{@{}c@{}}0.512\\ (0.501--0.524)\end{tabular} & \begin{tabular}[c]{@{}c@{}}0.503\\ (0.492--0.515)\end{tabular} & 23.626     & \textbf{0.172}     \\
    \begin{tabular}[c]{@{}c@{}}Transformer\\~\end{tabular}     & \begin{tabular}[c]{@{}c@{}}0.743\\(0.733--0.752)\end{tabular} & \begin{tabular}[c]{@{}c@{}}0.568\\ (0.562--0.574)\end{tabular} & \begin{tabular}[c]{@{}c@{}}0.543\\ (0.519--0.567)\end{tabular} & 419.680    & 1.380     \\
    \begin{tabular}[c]{@{}c@{}}LSTM\\~\end{tabular}    & \begin{tabular}[c]{@{}c@{}}0.719\\(0.706--0.732)\end{tabular} & \begin{tabular}[c]{@{}c@{}}0.468\\ (0.453--0.484)\end{tabular} & \begin{tabular}[c]{@{}c@{}}0.418\\ (0.381--0.454)\end{tabular} & 698.420    & 2.310     \\ \bottomrule
    \end{tabular}
\end{adjustbox}    
\end{table}

\section{Performance Analysis and Discussion}
\subsection{IBD Detection Performance}
Table~\ref{tab:performance} compares the testing performance and computational complexity of the final models in detecting the risk of IBD with a 1-month lead time. As we can see, GraD-IBD achieved a mean (95\% CI) testing AUROC, AP, and F1 score of 0.752 (0.745--0.759), 0.597 (0.591--0.604), and 0.612 (0.604--0.620), respectively, which outperformed all graph and sequential modeling methods. With ICD-graph as input, all graph models demonstrated significant reductions in both model size and computational cost compared to widely adopted sequential modeling approaches, without compromising performance. In fact, all graph-based models outperformed the classic LSTM model, which exhibited the highest computational cost. Moreover, as GraD-IBD can be seen as a context- and time-aware, directed variant of the GraphSAGE model, these two models achieved the same level of test AUROC; however, GraD-IBD still performed better than GraphSAGE in AP (0.597 vs. 0.582) and F1 score (0.612 vs. 0.565).

\begin{table}[!t]
\centering
\caption{Ablation study of context-aware and time-decay message passing mechanism in 10-fold cross-validation.}
\label{tab:ablation}
\begin{adjustbox}{width=0.45\textwidth}    \begin{threeparttable}
        \begin{tabular}{@{}cccc@{}}     
        \toprule
        \textbf{Information} & \textbf{AUROC\tnote{a}}     & \textbf{AP}   & \textbf{F1 score}     \\ \midrule
        \begin{tabular}[c]{@{}c@{}}GraD-IBD (CS+CF+TD\tnote{b} )\\~\end{tabular}   & \begin{tabular}[c]{@{}c@{}}\textbf{0.772}\\ (0.745--0.799)\end{tabular} & \begin{tabular}[c]{@{}c@{}}\textbf{0.598}\\ (0.554--0.641)\end{tabular} & \begin{tabular}[c]{@{}c@{}}\textbf{0.584}\\ (0.544--0.623)\end{tabular} \\
        \begin{tabular}[c]{@{}c@{}}CS+CF\\~\end{tabular} & \begin{tabular}[c]{@{}c@{}}0.765\\ (0.739--0.791)\end{tabular} & \begin{tabular}[c]{@{}c@{}}0.595\\ (0.555--0.636)\end{tabular} & \begin{tabular}[c]{@{}c@{}}0.582\\ (0.542--0.622)\end{tabular} \\
        \begin{tabular}[c]{@{}c@{}}CS+TD\\~\end{tabular} & \begin{tabular}[c]{@{}c@{}}0.761 \\ (0.740--0.782)\end{tabular} & \begin{tabular}[c]{@{}c@{}}0.590\\ (0.555--0.626)\end{tabular} & \begin{tabular}[c]{@{}c@{}}0.577\\ (0.544--0.610)\end{tabular}    \\
        \begin{tabular}[c]{@{}c@{}}CF+TD\\~\end{tabular} & \begin{tabular}[c]{@{}c@{}}0.756\\ (0.730--0.782)\end{tabular}  & \begin{tabular}[c]{@{}c@{}}0.593\\ (0.553--0.633)\end{tabular} & \begin{tabular}[c]{@{}c@{}}0.582\\ (0.541--0.623)\end{tabular}    \\
        \begin{tabular}[c]{@{}c@{}}TD\\~\end{tabular} & \begin{tabular}[c]{@{}c@{}}0.753\\ (0.724--0.781)\end{tabular}  & \begin{tabular}[c]{@{}c@{}}0.586\\ (0.545--0.627)\end{tabular} & \begin{tabular}[c]{@{}c@{}}0.575\\ (0.529--0.622)\end{tabular}    \\
        \begin{tabular}[c]{@{}c@{}}Uniform\\~\end{tabular} & \begin{tabular}[c]{@{}c@{}}0.751\\ (0.726--0.777)\end{tabular}  & \begin{tabular}[c]{@{}c@{}}0.558\\ (0.522--0.593)\end{tabular} & \begin{tabular}[c]{@{}c@{}}0.565\\ (0.531--0.600)\end{tabular} \\ \bottomrule
        \end{tabular}
        \begin{tablenotes}
            \item[\tnote{a}] All metrics were obtained from the cross-validation phase.  
            \item[\tnote{b}] Information used for message passing, aggregation, and feature update. \\CS: code similarity, CF: code occurrence frequency, TD: time decay, Uniform: all edge weights equal 1, which is equivalent to GraphSAGE.
        \end{tablenotes}
    \end{threeparttable}
    \end{adjustbox}
\vspace{-4mm}
\end{table}

\begin{figure*}[!t]
    \centering

    \makebox[\textwidth][c]{%
        \subfloat[AUROC]{%
            \includegraphics[width=0.333\textwidth]{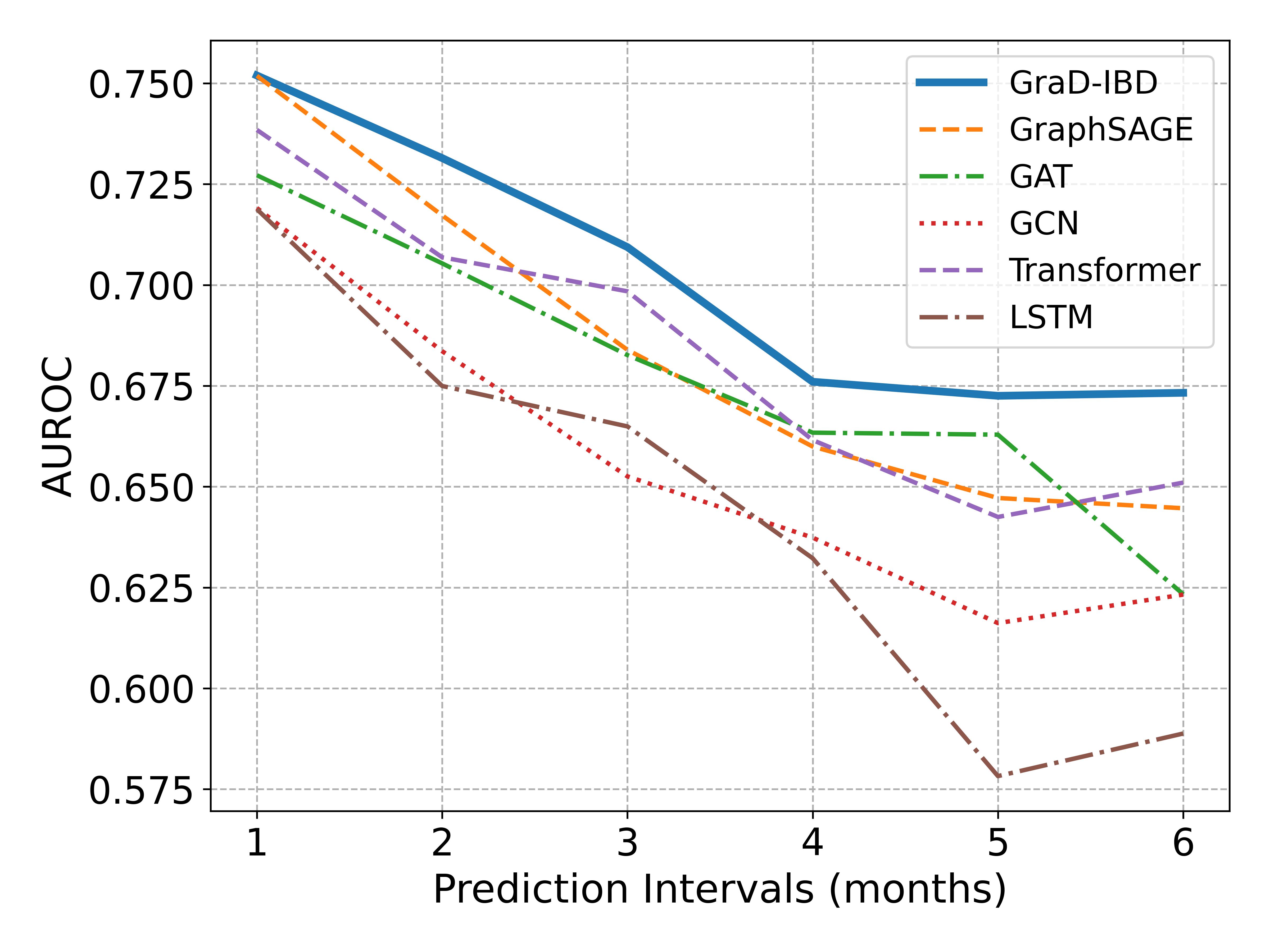}%
        }%
        \hspace{0.01\textwidth}%
        \subfloat[AP]{%
            \includegraphics[width=0.333\textwidth]{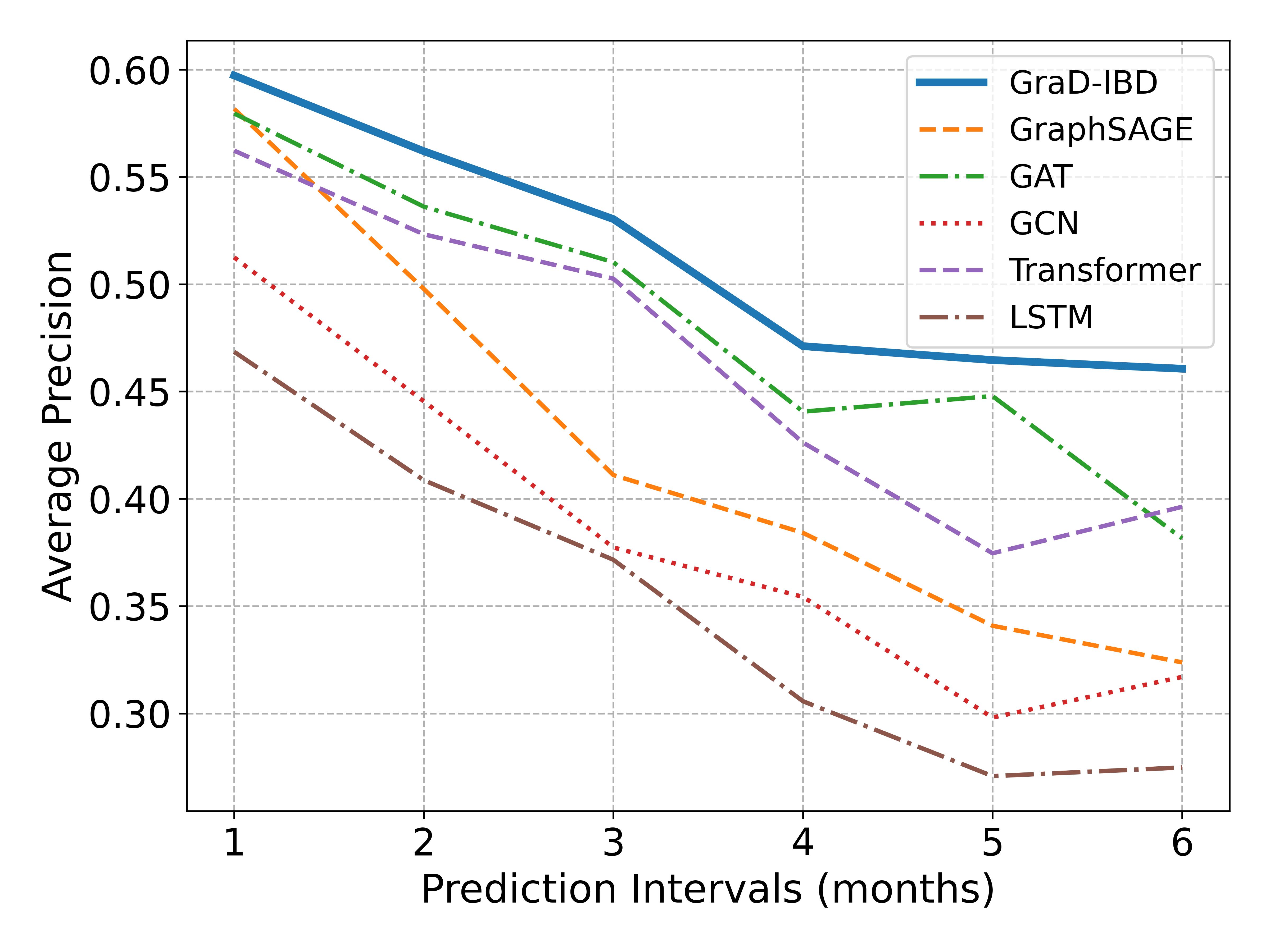}%
        }%
        \hspace{0.01\textwidth}%
        \subfloat[F1 score]{%
            \includegraphics[width=0.333\textwidth]{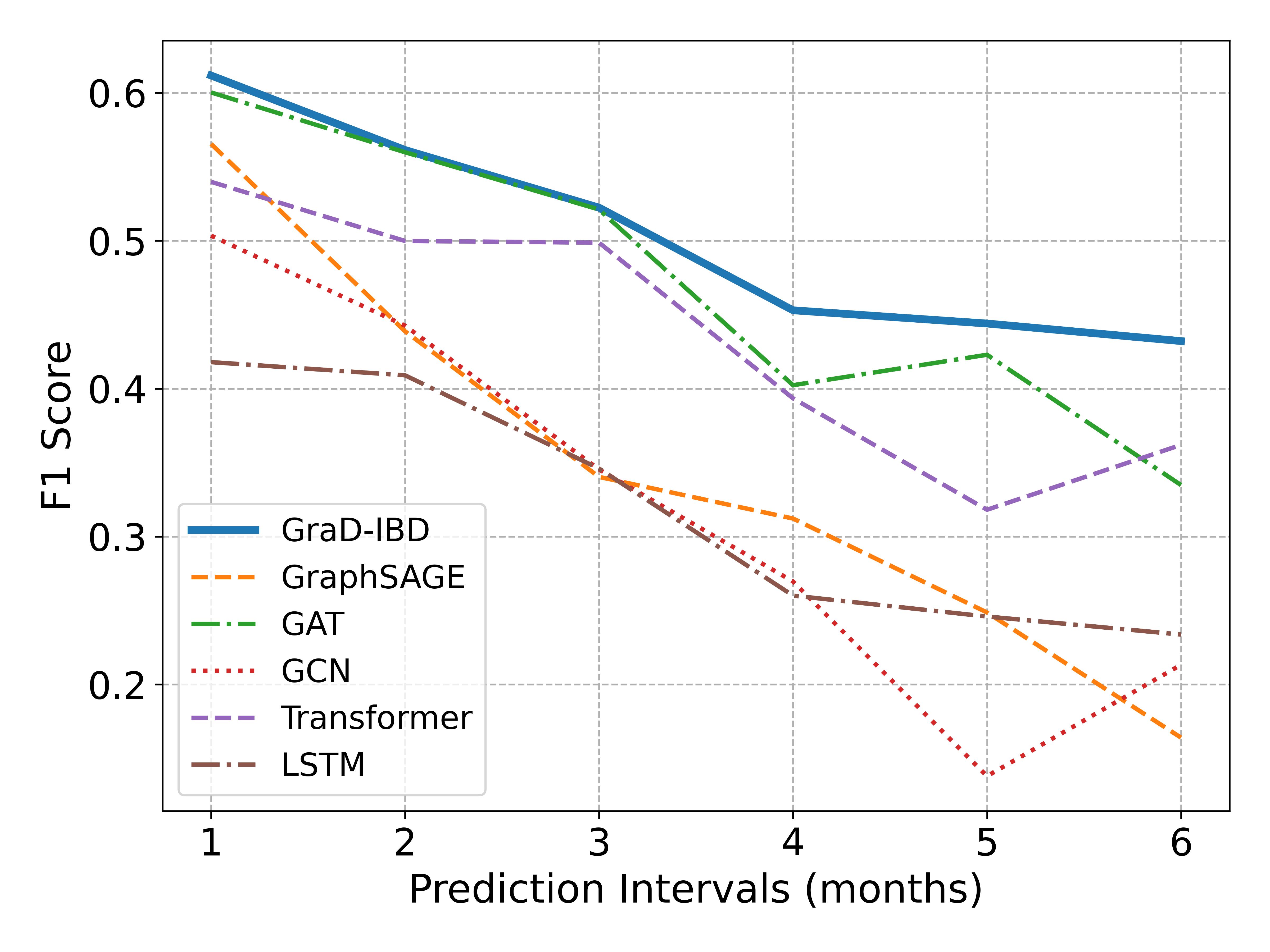}%
        }%
    }
    
    \caption{Testing performance of different graph and sequential modeling methods with variable prediction intervals ranging from 1 to 6 months: (a) AUROC, (b) AP, and (c) F1 score (obtained with a decision threshold of 0.5). }
    \label{fig:perf_intervals}
\end{figure*}

\subsection{Ablation Study}
During model tuning via cross-validation, we also conducted ablation experiments to assess the effectiveness of the proposed context-aware, time-decay message-passing mechanism. Table~\ref{tab:ablation} presents the results from 10-fold CV for different information combinations used in message passing when predicting the onset risk of IBD in 1 month into the future. As shown, when using both code similarity and occurrence frequency in edge weights, along with time decay in the feature update, our GraD-IBD model performed best across all three metrics compared to other message-passing strategies (row 1). Furthermore, across different parameter combinations (rows 2--4), each factor appeared to make distinct contributions to performance. By comparison, when the model did not account for either contextual or temporal associations, i.e., the uniform-weight case in row 6, it performed worst across all selections. As the uniform-weight case is also equivalent to the GraphSAGE model, this emphasizes the efficacy of context-aware, time-decay message passing, which contributes to the major improvement of the proposed method.

Additionally, it is worth noting again that the results of the ablation study shown in Table~\ref{tab:ablation} were obtained from the cross-validation phase, which aimed to find the optimal parameters, whereas Table~\ref{tab:performance} presents the testing performance with the final models. This experimental setting ensures test independence and objectivity.

\subsection{Sensitivity Analysis}
Lastly, we performed a sensitivity analysis to assess the robustness of the GraD-IBD model across tasks with different prediction intervals. This aims to evaluate the model's performance in longer-term risk triage to meet the needs for different clinical application scenarios. Fig.~\ref{fig:perf_intervals} illustrates the mean AUROC, AP, and F1 score as a function of the prediction interval ranging from 1 to 6 months. As expected, the task became increasingly challenging with longer lead times, resulting in lower performance across all models. Nevertheless, GraD-IBD consistently achieved superior performance metrics across all baseline models in all tasks. Notably, the performance advantage of our model became increasingly pronounced and robust as the prediction interval lengthened, such as over 3 months ahead. For instance, Grad-IBD outperformed the GraphSAGE model by $24\%$, $35\%$, and $44\%$ in terms of average precision for 4-, 5-, and 6-month prediction intervals, respectively. Meanwhile, the gap in AP between GraD-IBD and GraphSAGE was $3\%$ with the 1-month prediction interval.

\subsection{Limitation and Future Directions}
The model was developed using data from multiple hospitals within the same healthcare system. The institutional diagnosis pattern and coding conventions may potentially introduce bias into the model's decision-making process. As such, it would be highly valuable to include external validation datasets from other centers, as well as to extend it to other disease detection tasks, for a comprehensive assessment of its generalizability. Furthermore, investigations into model interpretability would be another essential aspect for better understanding the model's decisions and providing further insights into its credibility and clinical utility.

\section{Conclusion}
In this work, we present an efficient and powerful method for early detection of IBD by formatting irregular ICD sequences as patient-level, visit-bucketized, temporally directed graphs. Our experimental results demonstrate that the proposed GraD-IBD model outperforms various state-of-the-art graph and sequence modeling methods, achieving lower computational costs and, therefore, holding potential for a broader range of application scenarios. 

\section{Ethics Statement}
This study was approved by the Institutional Review Board of Mayo Clinic (24-009755) and conducted in accordance with applicable ethical guidelines and regulations.

\bibliographystyle{IEEEtran}
\bibliography{refs}

@misc{world2009international,
  title = {International Classification of Diseases ({ICD})},
  author= {{World Health Organization}},
  url = {https://www.who.int/standards/classifications/classification-of-diseases},
  note = {Accessed: 2025-07-25}
}

@inproceedings{choudhary2024graph,
  title={Graph Representation of Postoperative Patients for Opioids Refill Prediction: A Real-World Case Study},
  author={Choudhary, Ashok and Thiels, Cornelius A and Salehinejad, Hojjat},
  booktitle={2024 46th Annual International Conference of the IEEE Engineering in Medicine and Biology Society (EMBC)},
  pages={1--4},
  year={2024},
  organization={IEEE}
}

@article{li2020behrt,
  title={{BEHRT}: transformer for electronic health records},
  author={Li, Yikuan and Rao, Shishir and Solares, Jos{\'e} Roberto Ayala and Hassaine, Abdelaali and Ramakrishnan, Rema and Canoy, Dexter and Zhu, Yajie and Rahimi, Kazem and Salimi-Khorshidi, Gholamreza},
  journal={Scientific reports},
  volume={10},
  number={1},
  pages={7155},
  year={2020},
  publisher={Nature Publishing Group UK London}
}

@article{salehinejad2023contrastive,
  title={Contrastive transfer learning for prediction of adverse events in hospitalized patients},
  author={Salehinejad, Hojjat and Meehan, Anne M and Caraballo, Pedro J and Borah, Bijan J},
  journal={IEEE Journal of Translational Engineering in Health and Medicine},
  volume={12},
  pages={215--224},
  year={2023},
  publisher={IEEE}
}

@inproceedings{salehinejad2025deep,
  title={Deep Learning on {Hester Davis} Scores for Inpatient Fall Prediction},
  author={Salehinejad, Hojjat and Rojas, Ricky and Iheasirim, Kingsley and Yousufuddin, Mohammed and Borah, Bijan},
  booktitle={2025 IEEE Symposium on Computational Intelligence in Health and Medicine (CIHM)},
  pages={1--6},
  year={2025},
  organization={IEEE}
}

@article{salehinejad2023novel,
  title={Novel machine learning model to improve performance of an early warning system in hospitalized patients: a retrospective multisite cross-validation study},
  author={Salehinejad, Hojjat and Meehan, Anne M and Rahman, Parvez A and Core, Marcia A and Borah, Bijan J and Caraballo, Pedro J},
  journal={EClinicalMedicine},
  volume={66},
  year={2023},
  publisher={Elsevier}
}

@article{rajkomar2018scalable,
  title={Scalable and accurate deep learning with electronic health records},
  author={Rajkomar, Alvin and Oren, Eyal and Chen, Kai and Dai, Andrew M and Hajaj, Nissan and Hardt, Michaela and Liu, Peter J and Liu, Xiaobing and Marcus, Jake and Sun, Mimi and others},
  journal={NPJ digital medicine},
  volume={1},
  number={1},
  pages={18},
  year={2018},
  publisher={Nature Publishing Group UK London}
}

@article{junyuan2019pre,
  title={Pre-training of graph augmented transformers for medication recommendation},
  author={Junyuan, Shang and Tengfei, Ma and Cao, Xiao and Jimeng, Sun},
  journal={arXiv preprint arXiv:1906.00346},
  year={2019}
}

@inproceedings{choi2017gram,
  title={{GRAM}: graph-based attention model for healthcare representation learning},
  author={Choi, Edward and Bahadori, Mohammad Taha and Song, Le and Stewart, Walter F and Sun, Jimeng},
  booktitle={Proceedings of the 23rd ACM SIGKDD international conference on knowledge discovery and data mining},
  pages={787--795},
  year={2017}
}

@article{jiang2023graphcare,
  title={{Graphcare}: Enhancing healthcare predictions with personalized knowledge graphs},
  author={Jiang, Pengcheng and Xiao, Cao and Cross, Adam and Sun, Jimeng},
  journal={arXiv preprint arXiv:2305.12788},
  year={2023}
}

@inproceedings{choi2020learning,
  title={Learning the graphical structure of electronic health records with graph convolutional transformer},
  author={Choi, Edward and Xu, Zhen and Li, Yujia and Dusenberry, Michael and Flores, Gerardo and Xue, Emily and Dai, Andrew},
  booktitle={Proceedings of the AAAI conference on artificial intelligence},
  volume={34},
  pages={606--613},
  year={2020}
}

@inproceedings{wen2022disentangled,
  title={Disentangled dynamic heterogeneous graph learning for opioid overdose prediction},
  author={Wen, Qianlong and Ouyang, Zhongyu and Zhang, Jianfei and Qian, Yiyue and Ye, Yanfang and Zhang, Chuxu},
  booktitle={Proceedings of the 28th ACM SIGKDD Conference on Knowledge Discovery and Data Mining},
  pages={2009--2019},
  year={2022}
}

@inproceedings{chen2024predictive,
  title={Predictive modeling with temporal graphical representation on electronic health records},
  author={Chen, Jiayuan and Yin, Changchang and Wang, Yuanlong and Zhang, Ping},
  booktitle={IJCAI: proceedings of the conference},
  volume={2024},
  pages={5763},
  year={2024}
}

@article{baumgart2007inflammatory,
  title={Inflammatory bowel disease: cause and immunobiology},
  author={Baumgart, Daniel C and Carding, Simon R},
  journal={The Lancet},
  volume={369},
  number={9573},
  pages={1627--1640},
  year={2007},
  publisher={Elsevier}
}

@article{hamilton2017inductive,
  title={Inductive representation learning on large graphs},
  author={Hamilton, Will and Ying, Zhitao and Leskovec, Jure},
  journal={Advances in neural information processing systems},
  volume={30},
  year={2017}
}

@article{kingma2014adam,
  title={Adam: A method for stochastic optimization},
  author={Kingma, Diederik P and Ba, Jimmy},
  journal={arXiv preprint arXiv:1412.6980},
  year={2014}
}

@article{velivckovic2017graph,
  title={Graph attention networks},
  author={Veli{\v{c}}kovi{\'c}, Petar and Cucurull, Guillem and Casanova, Arantxa and Romero, Adriana and Lio, Pietro and Bengio, Yoshua},
  journal={arXiv preprint arXiv:1710.10903},
  year={2017}
}

@article{kipf2016semi,
  title={Semi-Supervised Classification with Graph Convolutional Networks},
  author={Kipf, Thomas N. and Welling, Max},
  journal={arXiv preprint arXiv:1609.02907},
  year={2016}
}

@article{vaswani2017attention,
  title={Attention is all you need},
  author={Vaswani, Ashish and Shazeer, Noam and Parmar, Niki and Uszkoreit, Jakob and Jones, Llion and Gomez, Aidan N and Kaiser, {\L}ukasz and Polosukhin, Illia},
  journal={Advances in neural information processing systems},
  volume={30},
  year={2017}
}

@article{hochreiter1997long,
  title={Long short-term memory},
  author={Hochreiter, Sepp and Schmidhuber, J{\"u}rgen},
  journal={Neural computation},
  volume={9},
  number={8},
  pages={1735--1780},
  year={1997},
  publisher={MIT press}
}

@article{devlin2018bert,
  title={{BERT}: Pre-training of Deep Bidirectional Transformers for Language Understanding},
  author={Devlin, Jacob and Chang, Ming-Wei and Lee, Kenton and Toutanova, Kristina},
  journal={arXiv preprint arXiv:1810.04805},
  year={2018}
}

@article{jayasooriya2023systematic,
  title={Systematic review with meta-analysis: time to diagnosis and the impact of delayed diagnosis on clinical outcomes in inflammatory bowel disease},
  author={Jayasooriya, Nishani and Baillie, Samantha and Blackwell, Jonathan and Bottle, Alex and Petersen, Irene and Creese, Hanna and Saxena, Sonia and Pollok, Richard C and POP-IBD study group},
  journal={Alimentary pharmacology \& therapeutics},
  volume={57},
  number={6},
  pages={635--652},
  year={2023},
  publisher={Wiley Online Library}
}

@misc{CDCstatsWellnessVisit2024, 
  author = {{National Center for Health Statistics}},
  title = {Percentage of having a wellness visit in past 12 months for adults aged 18 and over, {United States}, 2019—2024},
  year = {2024},
  url = {https://wwwn.cdc.gov/NHISDataQueryTool/SHS\_adult/index.html},
  note = {Accessed: 2025-07-20}
}

@misc{CDCNoInsurance2021, 
  author = {{Centers for Disease Control and Prevention}},
  title = {No health insurance coverage among people under age 65, by selected characteristics: {United States}, selected years 1984–2019},
  year = {2021},
  url={https://www.cdc.gov/nchs/data/hus/2020-2021/HINone.pdf},
  note = {Accessed: 2025-07-20}
}

@article{perler2019presenting,
  title={Presenting symptoms in inflammatory bowel disease: descriptive analysis of a community-based inception cohort},
  author={Perler, Bryce K and Ungaro, Ryan and Baird, Grayson and Mallette, Meaghan and Bright, Renee and Shah, Samir and Shapiro, Jason and Sands, Bruce E},
  journal={BMC gastroenterology},
  volume={19},
  number={1},
  pages={47},
  year={2019},
  publisher={Springer}
}

@article{lecun2015deep,
  title={Deep learning},
  author={LeCun, Yann and Bengio, Yoshua and Hinton, Geoffrey},
  journal={Nature},
  volume={521},
  number={7553},
  pages={436--444},
  year={2015},
  publisher={Nature Publishing Group UK London}
}

@article{ba2016layer,
  title={Layer normalization},
  author={Ba, Jimmy Lei and Kiros, Jamie Ryan and Hinton, Geoffrey E},
  journal={arXiv preprint arXiv:1607.06450},
  year={2016}
}

\end{document}